\documentclass{article}

\usepackage{amssymb}
\usepackage{amsmath}

\usepackage{microtype}
\usepackage{graphicx}
\usepackage{subfigure}
\usepackage{booktabs} 

\usepackage{hyperref}

\usepackage[accepted]{icml2018}

\icmltitlerunning{Generating Multi-Categorical Samples with Generative Adversarial Networks}

\begin{document}

\twocolumn[
\icmltitle{Generating Multi-Categorical Samples with Generative Adversarial Networks}



\icmlsetsymbol{equal}{*}

\begin{icmlauthorlist}
\icmlauthor{Ramiro D. Camino}{sedan}
\icmlauthor{Christian A. Hammerschmidt}{sedan}
\icmlauthor{Radu State}{sedan}
\end{icmlauthorlist}

\icmlaffiliation{sedan}{SEDAN Lab (SNT), University of Luxembourg, Luxembourg}

\icmlcorrespondingauthor{Ramiro D. Camino}{ramiro.camino@uni.lu}
\icmlcorrespondingauthor{Christian A. Hammerschmidt}{chrishammerschmidt@posteo.de}
\icmlcorrespondingauthor{Radu State}{radu.state@uni.lu}

\icmlkeywords{Machine Learning, Deep Learning, Neural Networks, Generative Models, Generative Adversarial Networks, Discrete Distributions, Categorical Variables}

\vskip 0.3in
]



\begin{NoHyper}
\printAffiliationsAndNotice{}  
\end{NoHyper}

\begin{abstract}
We propose a method to train generative adversarial networks on mutivariate feature vectors representing multiple categorical values.
In contrast to the continuous domain, where GAN-based methods have delivered considerable results, GANs struggle to perform equally well on discrete data.
We propose and compare several architectures based on multiple (Gumbel) softmax output layers taking into account the structure of the data.
We evaluate the performance of our architecture on datasets with different sparsity, number of features, ranges of categorical values, and dependencies among the features.
Our proposed architecture and method outperforms existing models.
\end{abstract}
\section{Introduction}
\label{introduction}

With recent advances in the field of generative adversarial networks (GANs)~\cite{goodfellow_generative_2014}, such as learning to transfer properties~\cite{kim_learning_2017} or understanding practical aspects \cite{gulrajani_improved_2017} as well as theoretical aspects \cite{mescheder_which_2018}, it is tempting to apply GANs to problems in data science tasks.
Most popular studies related to GANs involve continuous datasets from the computer vision domain, but data science applications usually deal with multiple continuous and categorical variables.
Training networks with categorical outputs raises additional challenges, because layers sampling from discrete distributions are often non-differentiable, which makes it impossible to train the network using backpropagation.
Existing models that work with discrete data are focused on generating sequential samples from a single categorical variable or vocabulary \cite{kusner_gans_2016,yu_seqgan:_2017,junbo_adversarially_2017,gulrajani_improved_2017}. 
In \cite{choi_generating_2017} multiple categorical variables are interpreted as a flat collection of binary features resulting from multiple one-hot-encodings, discarding useful information about the structure of the data.
To the extent of our knowledge, this is the first work that takes into account use cases where the data is composed by multiple categorical variables.
The specific contributions of this work are the following:
1) We present a review of the state of the art in generating discrete samples using GANs;
2) We propose a new model architecture with the corresponding training loss modifications by adapting the existing methods to the multi-categorical setting;
3) We extend the evaluation metrics used in literature to the multi-categorical setting and we show empirically that our approach outperforms the existing methods.

\section{Related Work}
\label{related}

While none of the existing work addresses learning from distributions with multiple categorical variables, different aspects of this problem have been addressed previously.  %

The output of a neural network can be transformed into a multinominal distribution by passing it through a softmax layer. However, sampling from this distribution is not a differentiable operation, which blocks the backpropagation process during the training of generative models for discrete samples. The Gumbel-Softmax \cite{jang_categorical_2016} and the Concrete-Distribution \cite{maddison_concrete_2016} were simultaneously proposed to tackle this problem in the domain of variational autoencoders (VAE) \cite{kingma_auto-encoding_2013}. Later \cite{kusner_gans_2016} adapted the technique to GANs for sequences of discrete elements.

Addressing the same problem, a reinforcement learning approach called SeqGAN \cite{yu_seqgan:_2017} interprets the generator as a stochastic policy and performs gradient policy updates to avoid the problem of backpropagation with discrete sequences. The discriminator in this case outputs the reinforcement learning reward for a full sequence, and several simulations generated with Monte Carlo search are executed to complete the missing steps.

An alternative approach that avoids backpropagating through discrete samples, called Adversarially regularized autoencoders (ARAE) \cite{junbo_adversarially_2017}, transforms sequences from a discrete vocabulary into a continuous latent space while simultaneously training both a generator (to output samples from the same latent distribution) and a discriminator (to distinguish between real and fake latent codes). Instead of using the traditional GAN architecture, the authors use the Wasserstein GAN (WGAN) \cite{arjovsky_wasserstein_2017} to improve training stability and obtain a loss more correlated with sample quality.

An architecture comparable to ARAE is presented as MedGAN \cite{choi_generating_2017} to synthesize realistic health care patient records.
The authors analyze the problem of generating simultaneously discrete and continuous samples while representing records as a mix of binary and numeric features.
The method pre-trains an autoencoder and then the generator returns latent codes as in the previous case, but they pass that output to the decoder before sending it to the discriminator; therefore, the discriminator receives either fake or real samples directly instead of continuous codes.
They propose using shortcut connections in the generator and a new technique which they refer to as minibatch averaging.
In order to better evaluate the generation of a whole batch (instead of individual isolated samples), minibatch averaging appends the average value per dimension of a batch to the batch itself before feeding it to the discriminator.

To address the difficulty of training GANs, an improved version for WGAN is presented in \cite{gulrajani_improved_2017} adding a gradient penalty to the critic loss (WGAN-GP) and removing the size limitation of the critic parameters. The authors present several use cases; in particular for the generation of word sequences, they claim that with this method discrete samples can be generated just by passing the outputs of softmax layers from the generator to the discriminator without sampling from them during training.

\section{Approaches to Multi-Categorical GANs}
\label{approach}
We compare in total six different networks: first the two baseline approaches, ARAE and MedGAN, with minimal adaptations to our use case.
Second, in order to improve the performance, we propose four variants based on an architecture with a modified generator for Gumbel-Softmax GAN and WGAN-GP, as well as using a modified decoder for ARAE and MedGAN.
Section~\ref{definitions} introduces the required definition and references. Sections \ref{multi-categorical-setting} to \ref{multi-categorical-autoencoders} describe our contribution and the scenario it is used in.

\subsection{Preliminaries}
\label{definitions}

GANs \cite{goodfellow_generative_2014} approximate drawing samples from a true data distribution $\mathbb{P}_{data}$ using a parametrized deterministic generator $G(z; \theta_g)$ with $z$ drawn from a random prior $\mathbb{P}_z$, obtaining the generated data distribution $\mathbb{P}_{g}$. A parametrized discriminator $D(x; \theta_d)$ is trained to maximize the probability of assigning the correct real or fake label to $x$ while at the same time $G$ learns how to fool $D$. This adversarial setting can be thought as a two-player minimax game with the following value function:

\[\min_{\theta_g} \max_{\theta_d}
\mathbb{E}_{x \sim \mathbb{P}_{data}} \left[D(x)\right]
+
\mathbb{E}_{z \sim \mathbb{P}_z} \left[1 - D(G(z))\right]
\]

Before each generator update step, if the discriminator is trained until convergence, then minimizing the value function is equivalent to minimizing the Jensen-Shannon divergence between $\mathbb{P}_{data}$ and $\mathbb{P}_{g}$, but that lead to vanishing gradients.
In practice, the model is trained by alternating discriminator and generator learning steps with minibatches of size $m$ optimizing a separate loss for each case:

\[\def\arraystretch{0.5}\begin{array}{l}
\mathcal{L}_d = \frac{1}{m} \sum_{i=1}^m \log(D(x_i)) + \log(1 - D(G(z_i))) \\
\\
\mathcal{L}_g = \frac{1}{m} \sum_{i=1}^m \log(D(G(z_i)))
\end{array}
\]

Let $g_1, \ldots, g_d$ be i.i.d samples drawn from $Gumbel(0, 1) = -\log(-\log(u_i))$ with $u_i \sim U(0, 1)$. The Gumbel-Softmax can generate sample vectors $x \in \{0, 1\}^d$ based on inputs $a \in R^d$ (that can be the output of previous layers) and a temperature hyperparameter $\tau \in (0, \infty)$ by the formula:

\[\begin{array}{rl}
x_i = \dfrac{exp((\log(a_i) + g_i) / \tau)}{\sum_{j=1}^d exp((\log(a_j) + g_j) / \tau)} & i=1, \ldots, d 
\end{array}
\]

In \cite{kusner_gans_2016} both the generator and the discriminator are implemented using recurrent neural networks. The generator uses the Gumbel-Softmax output layer to generate sequences of discrete elements, and the discriminator decides if a sequences are real or fake.

In WGAN \cite{arjovsky_wasserstein_2017} the authors claim that the divergences minimized in GANs lead to training difficulties because they are potentially not continuous. They propose to use the Wasserstein-1 distance (also known as Earth-Mover) because it is continuous everywhere and differentiable almost everywhere. Also they show that the loss they propose is more correlated with sample quality. In practice, the difference is that the discriminator is replaced by a critic, which returns real values instead of binary outputs, and its parameters are limited to the range $\left[-c; c\right]$ for a small constant $c$. The losses for this method are defined as:

\[\def\arraystretch{0.5}\begin{array}{l}
\mathcal{L}_d = \frac{1}{m} \sum_{i=1}^m - D(x_i) + D(G(z_i)) \\
\\
\mathcal{L}_g = \frac{1}{m} \sum_{i=1}^m - D(G(z_i))
\end{array}
\]

WGAN-GP \cite{gulrajani_improved_2017} also presents a recurrent neural network architecture for text generation, but does so with three modifications: first, it uses softmax without sampling, second, it uses a critic instead of a discriminator and third, it adds a gradient penalty to the critic loss:

\[
\lambda \mathbb{E}_{\hat{x} \sim \mathbb{P}_{\hat{x}}}
\left[(||\nabla_{\hat{x}} D(\hat{x})||_2 - 1)^2\right]
\]

\noindent where $\lambda$ is a new hyperparameter called penalty coefficient, and $\mathbb{P}_{\hat{x}}$ is a distribution defined by sampling uniformly along straight lines between pairs of samples from $\mathbb{P}_{data}$ and $\mathbb{P}_{g}$.

ARAE \cite{junbo_adversarially_2017} alternates training steps between an autoencoder defined by parametrized functions $Enc$ and $Dec$, a generator that samples continuous latent codes, and a critic that distinguishes between real and fake codes. The reconstruction loss of the autoencoder depends on the problem, e.g., cross entropy loss for one-hot encoded data. The generator loss is the same as in WGAN, and for the critic the loss is similar but includes the encoder in the equation:

\[
\mathcal{L}_d = \frac{1}{m} \sum_{i=1}^m - D(Enc(x_i)) + D(G(z_i))
\]

Notice that the encoder is also trained during the critic learning step, but the gradients passed to it are regularized.

MedGAN \cite{choi_generating_2017} also implements an autoencoder between the generator and the discriminator, but in this case the discriminator receives real data or fake data obtained from decoding fake codes. The autoencoder is first pre-trained separately using binary cross entropy as reconstruction for datasets with binary features:

\[\def\arraystretch{0.5}\begin{array}{c}
\mathcal{L}_{rec} = \sum_{i=1}^m x_i \log(\tilde{x}_i) + (1 - x_i) \log(1 - \tilde{x}_i) \\
 \\
\tilde{x}_i = Dec(Enc(x_i))
\end{array}
\]

During the main training phase the gradients flow from the discriminator to the decoder and afterwards to the generator. The traditional GAN loss is modified in the following way:

\[\def\arraystretch{0.5}\begin{array}{l}
\mathcal{L}_d = \frac{1}{m} \sum_{i=1}^m \log(D(x_i)) + \log(1 - D(Dec(G(z_i)))) \\
\\
\mathcal{L}_g = \frac{1}{m} \sum_{i=1}^m \log(D(Dec(G(z_i))))
\end{array}
\]

\subsection{Multi-Categorical Setting}
\label{multi-categorical-setting}

We define the space $\mathcal{V} = \mathcal{V}_1 \times \ldots \times \mathcal{V}_N$, where each $\mathcal{V}_i$ is a vocabulary of $|\mathcal{V}_i| = d_i$ elements. Then we define a multi categorical variable $v \in \mathcal{V}$ as the concatenation of $N$ categorical variables $v^i \in \mathcal{V}_i$. Each categorical variable $v^i$ can be represented with a one-hot encoded vector $x^i \in \{0, 1\}^{d_i}$, where the dimension of $x^i$ corresponding to the value of $v^i$ is set to one and the rest of the dimensions are set to zero. Hence, a multi-categorical variable $v$ can be represented with a multi-one-hot encoded vector $x \in \{0, 1\}^d$, where $x = \left[x^1; \ldots; x^N\right]$ and $d = \sum_{i=1}^N d_i$.

\begin{figure}[ht]
\vskip 0.2in
\begin{center}
\centerline{\includegraphics[width=\columnwidth]{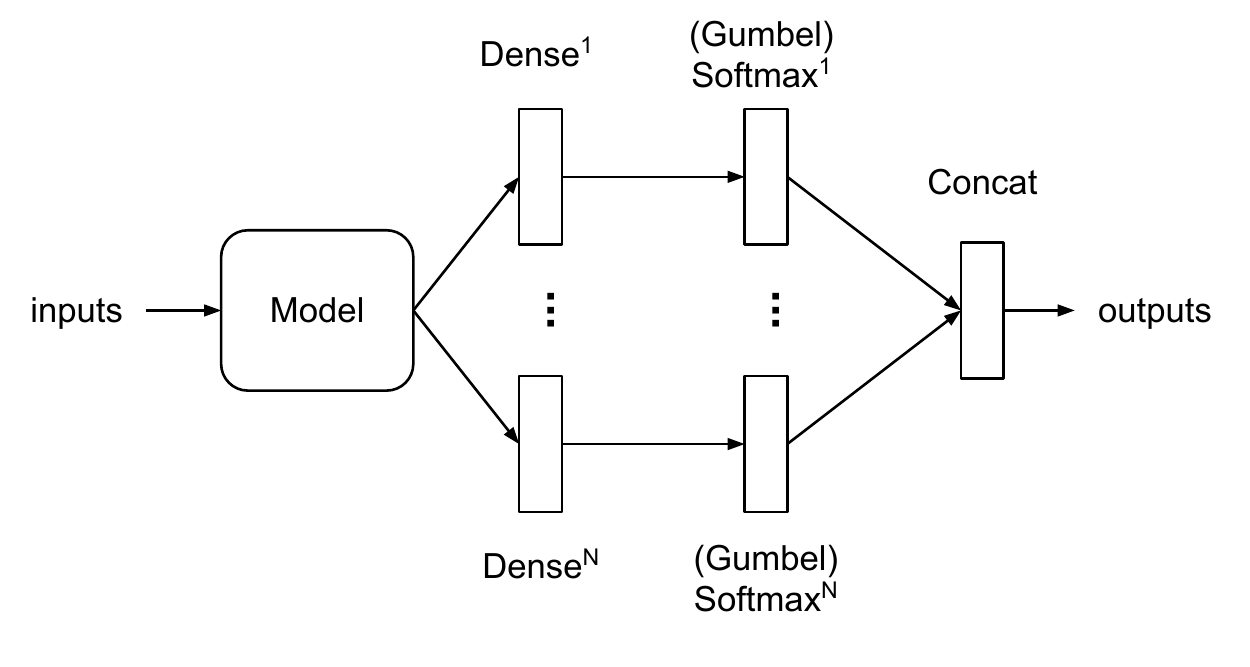}}
\caption{General architecture. After the model output, we place in parallel a dense layer per categorical variable, followed by an activation and a concatenation to obtain the final output. The choice between Gumbel-Softmax or Softmax depends on the model.}
\label{fig:architecture}
\end{center}
\vskip -0.2in
\end{figure}

Given a dataset $\mathcal{D} = \{x_1, \ldots, x_m\}$, where each element $x_i \in \mathcal{D}$ is a multi-one-hot encoding of $v_i \in \mathcal{V}$, our goal is to generate more samples following the same distribution. In order to do so, we propose in Figure \ref{fig:architecture} the architecture of the modified generator for Gumbel-Softmax GAN and WGAN-GP, as well as the modified decoder for ARAE and MedGAN.

\subsection{Multi-Categorical Generators}
\label{multi-categorical-generators}

In the case of Gumbel-Softmax GAN and WGAN-GP, we propose modifying the generator by splitting the output with a dense layer per categorical variable $\mathcal{V}_i$, each one transforming from the last hidden layer size to $d_i$. After that, we either apply a Gumbel-Softmax activation or in the case of WGAN-GP, simply a softmax activation. Finally, we concatenate the previous $N$ outputs into one final output of $d$ dimensions.

\subsection{Multi-Categorical Autoencoders}
\label{multi-categorical-autoencoders}

For ARAE and MedGAN, we propose modifying the decoder in a similar way, using in both cases a Gumbel-Softmax activation after splitting the output with a dense layer per categorical variable $\mathcal{V}_i$. The concatenation is the same in evaluation mode, but during training, we use the $N$ Gumbel-Softmax outputs separately to calculate a modified reconstruction loss that takes into account the multi-categorical structure. Using the notation $x^{j,k}$ to identify the $k$-th dimension of $x^j$ we define the reconstruction loss as:

\[
\mathcal{L}_{rec} =
\frac{1}{m} \sum_{i=1}^m \sum_{j=1}^N \sum_{k=1}^{d_j} -x_i^{j,k} \log \hat{x}_i^{j,k}
\]

\noindent which is the sum of the cross-entropy losses for each categorical variable between every sample and its reconstruction. 
\section{Experiments}
\label{sec:experiments}

To assess the quality of our proposal, we conduct experiments with varying data properties: sparsity, dimensionality, number of categories, and sample size.

\subsection{Datasets}
\label{subsec:datasets}

In order to have a better control of our experiments, we synthesize several datasets of one-hot encoded categorical variables. While the first categorical variable is uniformly distributed, the remaining ones were given a distribution at random based on the value of the previous variable. With this procedure we present four datasets of 10K samples: two with 10 variables of fixed size 2 and fixed size 10, one with 10 variables with random size between 2 and 10 and a bigger dataset of 100 variables with random size between 2 and 10. This setting allows us to evaluate the models with different configurations of sparsity and dimensionality. A summary of the four datasets is described in Table \ref{tab:datasets}.

\begin{table}[ht]
\caption{Datasets: number of samples, number of features and number of categorical variables with minimum and maximum size.}
\label{tab:datasets}
\vskip 0.15in
\begin{center}
\begin{small}
\begin{sc}
\begin{tabular}{lccccc}
\toprule
Name & $n$ & $d$ & $N$ & $d_{min}$ & $d_{max}$ \\
\midrule
Fixed 2 & 10K & 20 & 10 & 2 & 2 \\
Fixed 10 & 10K & 100 & 10 & 10 & 10 \\
Mix Small & 10K & 68 & 10 & 2 & 10 \\
Mix Big & 10K & 635 & 100 & 2 & 10 \\
Census & 2.5M & 396 & 68 & 2 & 18 \\
\bottomrule
\end{tabular}
\end{sc}
\end{small}
\end{center}
\vskip -0.1in
\end{table}

Additionally, we selected from the UCI Machine Learning Repository \cite{dua_uci_2017} the US Census Data (1990) Data Set to get a more realistic use case of multi-categorical variable data. It contains almost 2.5M samples with 68 variables, each of them with a size in the range from 2 to 18, transformed into 396 features.

\subsection{Evaluation methods}
\label{subsec:evaluation}

Evaluating the quality of GANs has proved to be difficult.
\citep{borji_pros_2018} provides a discussion of commonly used metrics. 
\citep{theis_note_2015} notes that not all measures provide a good assessment of the sample quality.

We partition each dataset $\mathcal{D}$ with a 90\%-10\% split into $\mathcal{D}_{train}$ and $\mathcal{D}_{test}$ in order to run our experiments. The models are trained with $\mathcal{D}_{train}$ to generate another dataset called $\mathcal{D}_{sample}$. The remaining $\mathcal{D}_{test}$ is used in different ways depending on the evaluation method.

The first two techniques are presented in \cite{choi_generating_2017} to perform a quantitative evaluation of the binary datasets. We start with the simplest one, which only checks if the model has learned the independent distribution of ones per dimension correctly. For datasets $\mathcal{D}_{sample}$ and $\mathcal{D}_{test}$, we calculate the frequency of ones $p(i)$ for all $1 \leq i \leq d$, obtaining two vectors $p_{sample}$ and $p_{test}$. Both are compared visually by the authors using scatter plots.

The problem with the initial approach is that it only captures how well the model can reproduce the independent feature distributions, but it does not take into account the dependencies between them. To tackle this, the second method creates one logistic regression model per feature $LR_{train_i}$, trained with $\mathcal{D}_{train}$ to predict feature $i$ using the rest of the features. Then, calculating the f1-score of each $LR_{train_i}$ using $\mathcal{D}_{test}$, we obtain the vector $f_{train}$. Repeating the same procedure but training with $\mathcal{D}_{sample}$ and evaluating with $\mathcal{D}_{test}$ we obtain $f_{sample}$. Both $f$ vectors are also compared visually by the authors using scatter plots.

During our experiments, we noticed that as the number of dimensions grows, the performance of the logistic regression models starts to deteriorate. 
This can be misleading, because if the f1 score is low, it is not very clear if it is caused by the properties of the dataset (either real or generated), or by the predictive power of the model. For that reason, we decided to replace logistic regression by random forests.

Furthermore, the second approach was crafted to evaluate binary samples. In our multi-categorical setting, given that we implement concatenated (Gumbel) softmax layers, each sample is represented as a multi-one-hot encoding, hence, only one dimension by categorical variable is set to one. 
This means that the task of predicting one dimension based on the remaining ones is almost trivial.
First, the missing dimension needs to be associated with the corresponding one-hot encoded categorical variable and then, there are only two cases: if there is a one in the one hot encoding, the missing dimension should be a zero, otherwise, it must be a one. 
Therefore, we propose a third method based on the second one, but predicting one categorical variable at a time instead of one feature at a time.
Each predictive model now handles a multi-class classification problem, hence we replace the f1-score by accuracy score. 
We collect two vectors $a_{train}$ and $a_{sample}$ that can be visually analyzed in the same way we mentioned for the previous vector pairs.

Finally we introduce a numerical analysis by measuring the mean squared error for each of the three vector metrics:

\[\def\arraystretch{1}\begin{array}{c}
MSE_p = MSE(p_{test}, p_{sample}) \\
 \\
MSE_f = MSE(f_{train}, f_{sample}) \\
 \\
MSE_a = MSE(a_{train}, a_{sample}) \\
\end{array}\]

\begin{figure*}[!ht]
\vskip 0.2in
\begin{center}
\centerline{\includegraphics[width=\textwidth]{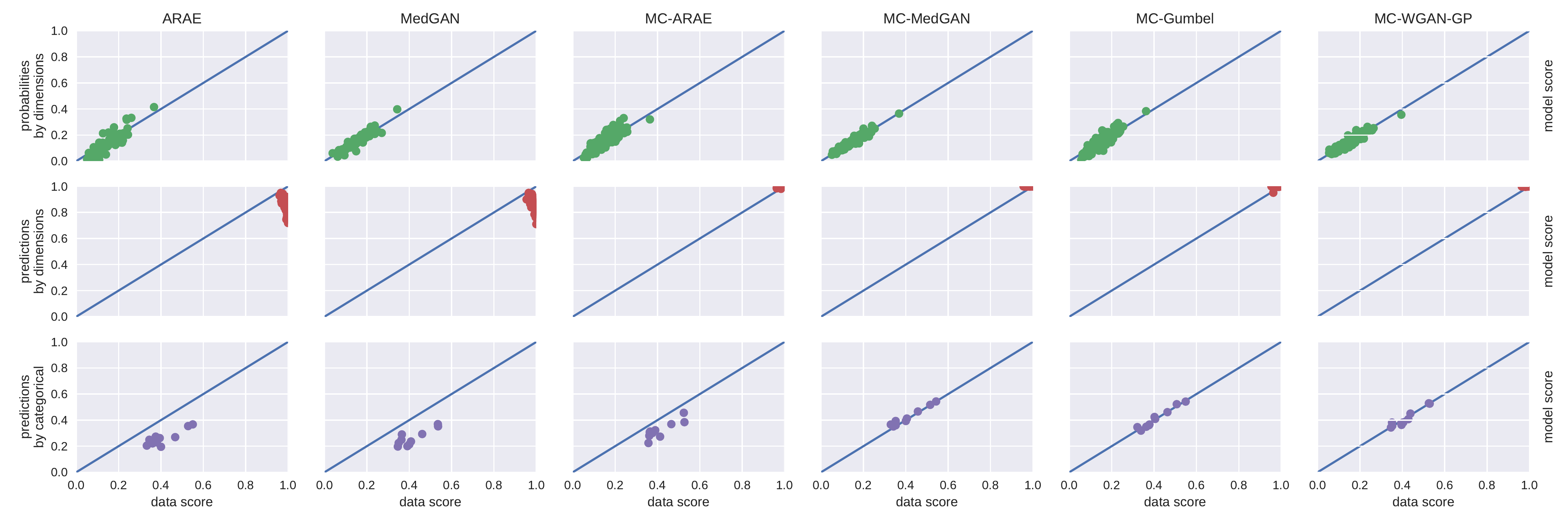}}
\caption{Scatter plots for a particular run on the \textsc{mix small} dataset, where every column is related to a model and every row to a metric. On the first row, the x-axis depicts the values of $p_{test}$ per dimension compared to the the values of $p_{sample}$ per dimension on the y-axis. The following two rows show analogue information comparing $f_{train}$ versus $f_{sample}$ and $a_{train}$ versus $a_{sample}$ respectively.}
\label{fig:plots}
\end{center}
\vskip -0.2in
\end{figure*}

Because of the randomness during training, we need to be sure that we did not get satisfactory results just by chance.
Thus, we train every model three times with different random seeds in order to capture the variance in these metrics.

\subsection{Implementation details}
\label{subsec:implementation-details}

We developed all our experiments using PyTorch 0.4.0
\footnote{Our open-source implementation is available at https://github.com/rcamino/multi-categorical-gans}
choosing hyperparameters with random search followed by manual tunning. In general, we trained using Adam \cite{kingma_adam:_2014} with minibatch size of 100 for 2000 epochs with learning rate $10^{-3}$. Compared to the other models, we had to experiment more to find the right setting for ARAE. We obtained better results by removing the original code normalization and noise annealing, training for more epochs with a smaller learning rate of $10^{-5}$.

Autoencoders were implemented with MLPs using tanh hidden activations. Generators and discriminators were implemented with MLPs with batch norm between layers (setting the decay to $0.99$), using ReLU activations for the generators and LeakyReLU for the discriminators. For ARAE we used the same amount of steps for the three training stages, and for the rest we used two steps for the discriminator for every step of the autoencoder and the generator.

In particular, we defined for ARAE the critic weights range to $\left[-0.1; 0.1\right]$. We used three hidden layers of size 100 for the generator and one for the discriminator, except for the MedGAN models where we used two hidden layers in the discriminator of sizes 256 and 128, and two hidden layers in the generator with the same size of the latent code because of the residual connections. For models with autoencoders, in the baseline cases the decoder used sigmoid outputs, and for the the multi-categorical versions we selected Gumbel-Softmax outputs with temperature $\tau=2/3$. Finally, for the simpler GAN with Gumbel-Softmax outputs the temperature was set to a smaller value of $\tau=1/3$.

\subsection{Results}
\label{subsec:results}

\begin{table*}[ht]
\vskip 0.15in
\begin{center}
\begin{small}
\begin{sc}
\begin{tabular}{lcccc}
\toprule
Method & Dataset & $MSE_p$ & $MSE_f$ & $MSE_a$ \\
\midrule
ARAE & FIXED 2 & 0.00031 $\pm$ 0.00004 & \textbf{0.00001} $\pm$ 0.00001 & 0.00059 $\pm$ 0.00022 \\
MedGAN & FIXED 2 & 0.00036 $\pm$ 0.00031 & 0.00005 $\pm$ 0.00003 & 0.00056 $\pm$ 0.00033 \\
MC-ARAE & FIXED 2 & 0.00046 $\pm$ 0.00028 & \textbf{0.00001} $\pm$ 0.00000 & 0.00058 $\pm$ 0.00024 \\
MC-MedGAN & FIXED 2 & \textbf{0.00013} $\pm$ 0.00006 & \textbf{0.00000} $\pm$ 0.00000 & \textbf{0.00032} $\pm$ 0.00017 \\
MC-GumbelGAN & FIXED 2 & 0.00337 $\pm$ 0.00188 & 0.00014 $\pm$ 0.00012 & 0.00050 $\pm$ 0.00012 \\
MC-WGAN-GP & FIXED 2 & 0.00030 $\pm$ 0.00007 & \textbf{0.00001} $\pm$ 0.00000 & 0.00068 $\pm$ 0.00012 \\
\midrule
ARAE & FIXED 10 & 0.00398 $\pm$ 0.00002 & 0.00274 $\pm$ 0.00021 & 0.02156 $\pm$ 0.00175 \\
MedGAN & FIXED 10 & 0.00720 $\pm$ 0.00825 & 0.00463 $\pm$ 0.00404 & 0.01961 $\pm$ 0.00214 \\
MC-ARAE & FIXED 10 & 0.00266 $\pm$ 0.00009 & \textbf{0.00036} $\pm$ 0.00018 & 0.01086 $\pm$ 0.00159 \\
MC-MedGAN & FIXED 10 & \textbf{0.00022} $\pm$ 0.00003 & 0.00167 $\pm$ 0.00010 & 0.00062 $\pm$ 0.00044 \\
MC-GumbelGAN & FIXED 10 & 0.00056 $\pm$ 0.00006 & 0.00110 $\pm$ 0.00013 & 0.00055 $\pm$ 0.00035 \\
MC-WGAN-GP & FIXED 10 & 0.00026 $\pm$ 0.00001 & 0.00123 $\pm$ 0.00005 & \textbf{0.00048} $\pm$ 0.00010 \\
\midrule
ARAE & MIX SMALL & 0.00261 $\pm$ 0.00020 & 0.01303 $\pm$ 0.00146 & 0.01560 $\pm$ 0.00039 \\
MedGAN & MIX SMALL & 0.00083 $\pm$ 0.00039 & 0.01889 $\pm$ 0.00258 & 0.02070 $\pm$ 0.00170 \\
MC-ARAE & MIX SMALL & 0.00195 $\pm$ 0.00040 & \textbf{0.00081} $\pm$ 0.00018 & 0.00759 $\pm$ 0.00100 \\
MC-MedGAN & MIX SMALL & \textbf{0.00029} $\pm$ 0.00003 & 0.00133 $\pm$ 0.00012 & 0.00080 $\pm$ 0.00018 \\
MC-GumbelGAN & MIX SMALL & 0.00078 $\pm$ 0.00027 & 0.00104 $\pm$ 0.00013 & 0.00047 $\pm$ 0.00008 \\
MC-WGAN-GP & MIX SMALL & 0.00048 $\pm$ 0.00010 & 0.00140 $\pm$ 0.00014 & \textbf{0.00037} $\pm$ 0.00016 \\
\midrule
ARAE & MIX BIG & 0.04209 $\pm$ 0.00362 & 0.02075 $\pm$ 0.01144 & 0.00519 $\pm$ 0.00087 \\
MedGAN & MIX BIG & 0.01023 $\pm$ 0.00263 & 0.00211 $\pm$ 0.00033 & 0.00708 $\pm$ 0.00162 \\
MC-ARAE & MIX BIG & 0.00800 $\pm$ 0.00019 & 0.00249 $\pm$ 0.00035 & 0.00472 $\pm$ 0.00092 \\
MC-MedGAN & MIX BIG & \textbf{0.00142} $\pm$ 0.00015 & 0.00491 $\pm$ 0.00055 & 0.01309 $\pm$ 0.00106 \\
MC-GumbelGAN & MIX BIG & 0.00312 $\pm$ 0.00032 & \textbf{0.00194} $\pm$ 0.00017 & \textbf{0.00430} $\pm$ 0.00021 \\
MC-WGAN-GP & MIX BIG & 0.00144 $\pm$ 0.00006 & 0.00536 $\pm$ 0.00030 & 0.01664 $\pm$ 0.00177 \\
\midrule
ARAE & CENSUS & 0.00165 $\pm$ 0.00082 & 0.00206 $\pm$ 0.00030 & 0.00668 $\pm$ 0.00175 \\
MedGAN & CENSUS & 0.00871 $\pm$ 0.01078 & 0.00709 $\pm$ 0.00889 & 0.01723 $\pm$ 0.02177 \\
MC-ARAE & CENSUS & 0.00333 $\pm$ 0.00020 & 0.00129 $\pm$ 0.00019 & 0.00360 $\pm$ 0.00095 \\
MC-MedGAN & CENSUS & \textbf{0.00012} $\pm$ 0.00004 & 0.00024 $\pm$ 0.00003 & 0.00013 $\pm$ 0.00003 \\
MC-GumbelGAN & CENSUS & 0.01866 $\pm$ 0.00040 & 0.00981 $\pm$ 0.00034 & 0.03930 $\pm$ 0.00469 \\
MC-WGAN-GP & CENSUS & 0.00019 $\pm$ 0.00004 & \textbf{0.00017} $\pm$ 0.00002 & \textbf{0.00008} $\pm$ 0.00002 \\
\bottomrule
\end{tabular}
\end{sc}
\end{small}
\end{center}
\caption{The three proposed metrics for every dataset and every model.}
\label{tab:results}
\vskip -0.1in
\end{table*}

In Table \ref{tab:results} we summarize the experiments we executed for every model on each dataset. For each of the three proposed metrics, we present the mean and the standard deviation of the values obtained in three separated runs using different random seeds. We make the following observations:

\begin{itemize}
\item The standard deviation for every case is usually smaller than 0.01, which indicates that there is not a significant difference in the metrics between runs and that the results are not a product of mere chance.

\item For the Fixed 2 dataset the improvement of the multi-categorical models is not evident, but it is reasonable given that samples with only two possible values per categorical variable are equivalent to binary samples, and the baseline models are prepared for that setting.

\item Using a collection of variables with the same size or with different sizes does not seem to affect the result.

\item The size and the sparsity of the samples seem to be the most important factors. When the datasets have more and larger variables, the dependencies between dimensions become more complex and also the independent proportion of ones per dimension are harder to learn. The experiments show that no model outperforms the rest for every case. It is reasonable to claim that for different settings a different model can be the best pick, but in general we can observe 
that the multi-categorical models give better results than the baselines.

\item We notice cases when, for example, Gumble-GAN is the best method in the \textsc{mix big} experiments, but the worst in the \textsc{Census} dataset.
Given that we deal with six models with a considerable amount of hyperparameters and architecture options, it might be valuable to do a more exhaustive hyperparameter search to identify the reason for the variance in performance.
\end{itemize}

As an example, we show in Figure~\ref{fig:plots} the scatter plots for the \textsc{mix small} dataset, that are associated with the values summarized in the third block of Table~\ref{tab:results}.
Because we plot ground-truth versus model output, perfect results should fall onto the diagonal line. We can see how the first row is fairly good for every model, and how the second row is almost perfect for our approach (as explained in \ref{subsec:evaluation}). Finally, for the last row, the last three columns corresponding to multi-categorical models present better performance over the rest.

\vfill\null

\section{Conclusion}
\label{conclusion}

We extended the autoencoder of MedGAN and ARAE as well as the generator of GumbleGAN and WGAN-GP with dense layers feeding into (Gumbel) softmax layers, each separated into the right dimensionality for their corresponding output variables.
Additionally, we adapted a metric to our categorical setting to compare all approaches.

Compared to unmodified MedGAN and ARAE, all approaches improve the performance across all datasets, even though we cannot identify a clear best model.
The performance improvement comes at the cost of requiring to add extra information, namely the dimensionality of the variables.
In some cases, this information may not be available.

Lastly, while we are positive about the usefulness of the synthetic data for our purposes, other applications might require different evaluation methods to determine the quality of the samples generated by each approach.
This might also guide the search for the right hyperparameters and architecture options.

\clearpage
\section*{Acknowledgements}

This project was funded by the \textit{University of Luxembourg}, the \textit{Fonds National de la Recherche (FNR)} and our industrial partner \textit{LOGOS IT Services SA}.

The experiments presented in this paper were carried out using the HPC facilities of the University of Luxembourg~\cite{VBCG_HPCS14} {\small -- see \url{https://hpc.uni.lu}}.
\bibliography{Zotero,HPC}

\begin{thebibliography}{17}
\providecommand{\natexlab}[1]{#1}
\providecommand{\url}[1]{\texttt{#1}}
\expandafter\ifx\csname urlstyle\endcsname\relax
  \providecommand{\doi}[1]{doi: #1}\else
  \providecommand{\doi}{doi: \begingroup \urlstyle{rm}\Url}\fi

\bibitem[Arjovsky et~al.(2017)Arjovsky, Chintala, and
  Bottou]{arjovsky_wasserstein_2017}
Arjovsky, Martin, Chintala, Soumith, and Bottou, Léon.
\newblock Wasserstein {GAN}.
\newblock \emph{arXiv preprint arXiv:1701.07875}, 2017.
\newblock URL \url{https://arxiv.org/abs/1701.07875}.

\bibitem[Borji(2018)]{borji_pros_2018}
Borji, Ali.
\newblock Pros and {Cons} of {GAN} {Evaluation} {Measures}.
\newblock \emph{arXiv:1802.03446 [cs]}, February 2018.
\newblock URL \url{http://arxiv.org/abs/1802.03446}.
\newblock arXiv: 1802.03446.

\bibitem[Choi et~al.(2017)Choi, Biswal, Malin, Duke, Stewart, and
  Sun]{choi_generating_2017}
Choi, Edward, Biswal, Siddharth, Malin, Bradley, Duke, Jon, Stewart, Walter~F.,
  and Sun, Jimeng.
\newblock Generating {Multi}-label {Discrete} {Patient} {Records} using
  {Generative} {Adversarial} {Networks}.
\newblock \emph{arXiv:1703.06490 [cs]}, March 2017.
\newblock URL \url{http://arxiv.org/abs/1703.06490}.
\newblock arXiv: 1703.06490.

\bibitem[Dua \& Taniskidou(2017)Dua and Taniskidou]{dua_uci_2017}
Dua, D. and Taniskidou, E.~Karra.
\newblock {UCI} {Machine} {Learning} {Repository}.
\newblock 2017.

\bibitem[Goodfellow et~al.(2014)Goodfellow, Pouget-Abadie, Mirza, Xu,
  Warde-Farley, Ozair, Courville, and Bengio]{goodfellow_generative_2014}
Goodfellow, Ian, Pouget-Abadie, Jean, Mirza, Mehdi, Xu, Bing, Warde-Farley,
  David, Ozair, Sherjil, Courville, Aaron, and Bengio, Yoshua.
\newblock Generative {Adversarial} {Nets}.
\newblock In Ghahramani, Z., Welling, M., Cortes, C., Lawrence, N.~D., and
  Weinberger, K.~Q. (eds.), \emph{Advances in {Neural} {Information}
  {Processing} {Systems} 27}, pp.\  2672--2680. Curran Associates, Inc., 2014.
\newblock URL
  \url{http://papers.nips.cc/paper/5423-generative-adversarial-nets.pdf}.

\bibitem[Gulrajani et~al.(2017)Gulrajani, Ahmed, Arjovsky, Dumoulin, and
  Courville]{gulrajani_improved_2017}
Gulrajani, Ishaan, Ahmed, Faruk, Arjovsky, Martin, Dumoulin, Vincent, and
  Courville, Aaron~C.
\newblock Improved training of wasserstein gans.
\newblock In \emph{Advances in {Neural} {Information} {Processing} {Systems}},
  pp.\  5769--5779, 2017.

\bibitem[Jang et~al.(2016)Jang, Gu, and Poole]{jang_categorical_2016}
Jang, Eric, Gu, Shixiang, and Poole, Ben.
\newblock Categorical {Reparameterization} with {Gumbel}-{Softmax}.
\newblock \emph{arXiv:1611.01144 [cs, stat]}, November 2016.
\newblock URL \url{http://arxiv.org/abs/1611.01144}.
\newblock arXiv: 1611.01144.

\bibitem[Junbo et~al.(2017)Junbo, Zhao, Kim, Zhang, Rush, and
  LeCun]{junbo_adversarially_2017}
Junbo, Zhao, Kim, Yoon, Zhang, Kelly, Rush, Alexander~M., and LeCun, Yann.
\newblock Adversarially {Regularized} {Autoencoders}.
\newblock \emph{arXiv:1706.04223 [cs]}, June 2017.
\newblock URL \url{http://arxiv.org/abs/1706.04223}.
\newblock arXiv: 1706.04223.

\bibitem[Kim et~al.(2017)Kim, Cha, Kim, Lee, and Kim]{kim_learning_2017}
Kim, Taeksoo, Cha, Moonsu, Kim, Hyunsoo, Lee, Jung~Kwon, and Kim, Jiwon.
\newblock Learning to {Discover} {Cross}-{Domain} {Relations} with {Generative}
  {Adversarial} {Networks}.
\newblock \emph{arXiv:1703.05192 [cs]}, March 2017.
\newblock URL \url{http://arxiv.org/abs/1703.05192}.
\newblock arXiv: 1703.05192.

\bibitem[Kingma \& Ba(2014)Kingma and Ba]{kingma_adam:_2014}
Kingma, Diederik~P. and Ba, Jimmy.
\newblock Adam: {A} method for stochastic optimization.
\newblock \emph{arXiv preprint arXiv:1412.6980}, 2014.

\bibitem[Kingma \& Welling(2013)Kingma and Welling]{kingma_auto-encoding_2013}
Kingma, Diederik~P. and Welling, Max.
\newblock Auto-{Encoding} {Variational} {Bayes}.
\newblock \emph{arXiv:1312.6114 [cs, stat]}, December 2013.
\newblock URL \url{http://arxiv.org/abs/1312.6114}.
\newblock arXiv: 1312.6114.

\bibitem[Kusner \& Hernández-Lobato(2016)Kusner and
  Hernández-Lobato]{kusner_gans_2016}
Kusner, Matt~J. and Hernández-Lobato, José~Miguel.
\newblock {GANS} for {Sequences} of {Discrete} {Elements} with the
  {Gumbel}-softmax {Distribution}.
\newblock \emph{arXiv:1611.04051 [cs, stat]}, November 2016.
\newblock URL \url{http://arxiv.org/abs/1611.04051}.
\newblock arXiv: 1611.04051.

\bibitem[Maddison et~al.(2016)Maddison, Mnih, and Teh]{maddison_concrete_2016}
Maddison, Chris~J., Mnih, Andriy, and Teh, Yee~Whye.
\newblock The {Concrete} {Distribution}: {A} {Continuous} {Relaxation} of
  {Discrete} {Random} {Variables}.
\newblock \emph{arXiv:1611.00712 [cs, stat]}, November 2016.
\newblock URL \url{http://arxiv.org/abs/1611.00712}.
\newblock arXiv: 1611.00712.

\bibitem[Mescheder et~al.(2018)Mescheder, Geiger, and
  Nowozin]{mescheder_which_2018}
Mescheder, Lars, Geiger, Andreas, and Nowozin, Sebastian.
\newblock Which {Training} {Methods} for {GANs} do actually {Converge}?
\newblock \emph{arXiv:1801.04406 [cs]}, January 2018.
\newblock URL \url{http://arxiv.org/abs/1801.04406}.
\newblock arXiv: 1801.04406.

\bibitem[Theis et~al.(2015)Theis, Oord, and Bethge]{theis_note_2015}
Theis, Lucas, Oord, Aäron van~den, and Bethge, Matthias.
\newblock A note on the evaluation of generative models.
\newblock \emph{arXiv:1511.01844 [cs, stat]}, November 2015.
\newblock URL \url{http://arxiv.org/abs/1511.01844}.
\newblock arXiv: 1511.01844.

\bibitem[Varrette et~al.(2014)Varrette, Bouvry, Cartiaux, and
  Georgatos]{VBCG_HPCS14}
Varrette, S., Bouvry, P., Cartiaux, H., and Georgatos, F.
\newblock Management of an academic hpc cluster: The ul experience.
\newblock In \emph{Proc. of the 2014 Intl. Conf. on High Performance Computing
  \& Simulation (HPCS 2014)}, pp.\  959--967, Bologna, Italy, July 2014. IEEE.

\bibitem[Yu et~al.(2017)Yu, Zhang, Wang, and Yu]{yu_seqgan:_2017}
Yu, Lantao, Zhang, Weinan, Wang, Jun, and Yu, Yong.
\newblock {SeqGAN}: {Sequence} {Generative} {Adversarial} {Nets} with {Policy}
  {Gradient}.
\newblock In \emph{{AAAI}}, pp.\  2852--2858, 2017.

\end{thebibliography}
\bibliographystyle{icml2018}

\end{document}